\documentclass[conference]{IEEEtran}
\IEEEoverridecommandlockouts
\usepackage{cite}
\usepackage{amsmath,amssymb,amsfonts}
\usepackage{algorithmic}
\usepackage{graphicx}
\usepackage{textcomp}
\usepackage{xcolor}
\def\BibTeX{{\rm B\kern-.05em{\sc i\kern-.025em b}\kern-.08em
    T\kern-.1667em\lower.7ex\hbox{E}\kern-.125emX}}

\usepackage{microtype}
\usepackage{graphicx}
\usepackage[caption=false]{subfig}
\usepackage{booktabs} 

\usepackage{hyperref}

\usepackage[utf8]{inputenc} 
\usepackage[T1]{fontenc}    
\usepackage{hyperref}       
\usepackage{url}            
\usepackage{booktabs}       
\usepackage{amsfonts}       
\usepackage{nicefrac}       
\usepackage{microtype}      

\usepackage{times}
\usepackage{epsfig}
\usepackage{graphicx}
\usepackage{amsmath}
\usepackage{amssymb}
\usepackage{xcolor}
\usepackage{dsfont}

\usepackage[utf8]{inputenc} 
\usepackage[T1]{fontenc}    
\usepackage{url}            
\usepackage{booktabs}       
\usepackage{amsfonts}       
\usepackage{nicefrac}       
\usepackage{microtype}      
\usepackage{graphicx}
\usepackage{mathabx}
\usepackage{longtable}
\usepackage{amsthm}   
\usepackage{stackrel}  

\usepackage{bm} 

\usepackage{float}  
\usepackage{multirow}
\usepackage{xcolor,colortbl} 
\usepackage{color}  
\usepackage[skip=2pt]{caption}  
\usepackage{enumitem}

\providecommand{\realnum}{\mathbb{R}}

\DeclareMathOperator*{\symm}{Symm}  
\providecommand{\citep}{\cite} 
\providecommand{\citet}{\cite}
\providecommand{\etal}{\textit{et~al.}}  
\DeclareMathOperator*{\myargmax}{arg\,max}
\providecommand{\modeldisent}{ST-GAN}
\providecommand{\modeldisentmain}{\modeldisent}
\providecommand{\modeltypedisent}{normalized independent component analysis model}

\begin{document}

\title{Unsupervised Controllable Generation \\with Self-Training}

\author{\IEEEauthorblockN{Grigorios G Chrysos$^*$\thanks{$^*$Work done during an internship at NVIDIA Research.}}
\IEEEauthorblockA{\textit{EPFL}\\
Switzerland \\
grigorios.chrysos@epfl.ch}
\and
\IEEEauthorblockN{Jean Kossaifi}
\IEEEauthorblockA{\textit{NVIDIA}\\
United States \\
jean.kossaifi@gmail.com}
\and
\IEEEauthorblockN{Zhiding Yu}
\IEEEauthorblockA{\textit{NVIDIA}\\
United States \\
zhidingy@nvidia.com}
\and
\IEEEauthorblockN{Anima Anandkumar}
\IEEEauthorblockA{\textit{Caltech / NVIDIA}\\
United States \\
aanandkumar@nvidia.com}
}

\maketitle

\begin{abstract}
Recent generative adversarial networks (GANs) are able to generate impressive photo-realistic images. However, controllable generation with GANs remains an open research problem. Achieving controllable generation requires semantically interpretable and  disentangled factors of variation. It is challenging to achieve this goal using simple fixed distributions such as Gaussian distribution. Instead, we propose an unsupervised framework to learn a distribution of latent codes that control the generator through self-training. Self-training provides an iterative feedback in the GAN training, from the discriminator to the generator, and progressively improves the proposal of the latent codes as training proceeds. The latent codes are sampled from a latent variable model that is learned in the feature space of the discriminator. We consider a \modeltypedisent{} and learn its parameters  through tensor factorization of the higher-order moments.
Our framework exhibits better disentanglement compared to other variants such as the variational autoencoder,  and is able to discover semantically meaningful latent codes without any supervision. We empirically demonstrate on both cars and faces datasets that each group of elements in the learned code controls a mode of variation with a semantic meaning, e.g. pose or background change. We also demonstrate with quantitative metrics that our method generates better results compared to other approaches. 
  
\end{abstract}
\section{Introduction}
\label{sec:disent_intro}

Generative Adversarial Networks (GANs)~\cite{goodfellow2014generative}
are the method of choice for synthesis,
owing to their ability to generate impressive photo-realistic images.
Yet, they fall short in a key aspect of generation in real-world applications, \emph{controllability} -- the ability to control 
the semantic of the generated images 
in an interpretable, deterministic manner.
Controllability will enable on-demand synthesis of images, which has numerous applications, including data augmentation and image editing.
Controllable generation relies on semantically interpretable disentangled factors of variation, i.e., factors that modify a single mode of variation, such as length or color of the hair.

However, the unsupervised nature of GANs
hinders the development of controllable generation.
For instance, given a generator of a standard GAN model trained on facial synthesis, 
it is not possible to directly control semantic attributes in a new synthesized instance,
such as type/length/color of the hair, shape of the face, etc. 
Adding supervision means getting access to various (labelled) image attributes, which can be expensive.
To reduce the amount of supervision required, 
Nie \etal~\cite{nie2020semi} consider semi-supervised learning in StyleGANs 
and reveal that limited amount of supervision is sufficient for high quality generation.
This assumes, nonetheless, that all the semantic attributes are labeled, 
which may not always be achievable.
Instead, it is preferable to achieve controlled 
generation in a fully unsupervised manner.

Unsupervised disentanglement has been explored in the Variational Autoencoders (VAEs)~\cite{kim2018disentangling,chen2018isolating}.
However, the generation quality of VAEs does not yet match the quality of GAN-synthesized images. 
GAN methods have also been extended 
to achieve disentanglement of the factors of variation~\cite{chen2016infogan,kaneko2018generative,lee2020high}. However, these unsupervised GAN and VAE approaches have two major drawbacks:
i) the disentangled factors are not guaranteed to be interpretable and 
ii) such models suffer from non-identifiability~\cite{locatello2018challenging}, meaning that different runs can produce different factors. 

Consequently, in this work, we propose \modeldisent, the first fully unsupervised approach to controllable generation with GANs through self-training. Specifically, \textbf{we make the following contributions}: 
\begin{itemize}[leftmargin=4.0mm]
    \item We propose a novel self-training procedure to discover disentangled and semantically interpretable latent codes driving the generation.
    The self training feedback loop allows for iterative refinement of the factor codes. We design a framework that encourages the model to produce interpretable factor codes
    that can faithfully control the generator.
    \item Instead of sampling from a fixed distribution probability distribution (e.g. Gaussian), we use a flexible latent variable model. Specifically, we employ a normalized independent component analysis models. To learn its parameters, we apply tensor factorization to the higher-order cumulants of the representation learnt in the feature space discriminator. We also experiment with a variational autoencoder as an alternative latent variable model. 
    \item We empirically demonstrate that our approach results in a controllable GAN able to learn the hidden codes in a fully unsupervised manner. We show on two different domains, cars and faces, that the discovered codes are disentangled and semantically interpretable, allowing to control the variations in the synthesized images. 
    \item We propose two quantitative metrics for measuring the semantic changes by modifying a single element of the factor codes. Using them, we quantitatively demonstrate significant improvements with our model over the baselines.Our experiments exhibit how our framework can progressively improve the learning of latent codes as the training proceeds. We also establish the importance of each block of our model through extensive ablation studies. 
\end{itemize}

 \section{Related work}
\label{sec:disent_related}

\paragraph{Controllable generation in GAN}: Both GAN and VAE models have been utilized for achieving controllability, while some studies even indicate that the techniques in one do not work in the other~\cite{lin2019infogan}. While the majority of existing work on disentanglement focuses on a (semi-)supervised setting~\cite{tran2017disentangled,zhang2019multi,xiao2017dna,reed2014learning,szabo2017challenges, deng2017structured,chongxuan2017triple,kossaifi2018gagan,tran2019disentangling}, our work focuses on the unsupervised setting. Here, we review the most closely related methods below. Such methods can be classified into three categories: a) post-training interpretable methods, b) information theory motivated disentanglement, c) hierarchical disentanglement. The post-processing methods~\cite{plumerault2020controlling, voynov2020unsupervised} assume a pretrained generator and find interpretable directions in the latent space of the generator. Post-processing methods have twofold drawbacks: i) they do not have any guarantees that we will discover such interpretable directions, ii) multi-step training is required.

The perspective of information theory is frequently used to tackle (unsupervised) disentanglement. The core idea relies on maximizing the mutual information between the latent codes and the synthesized images. The seminal work of InfoGAN~\cite{chen2016infogan} along with its extensions~\cite{lin2019infogan, liu2019oogan, jeon2018ib} use the mutual information to disentangle the factors of variation in a GAN setting. The drawback of mutual information is that they require strong inductive biases to disentangle the factors of variation~\cite{lin2019infogan}.

A line of work that relates to our method is that of hierarchical disentanglement. FineGAN~\cite{singh2019finegan} utilizes bounding boxes to define a hierarchical disentanglement. Kaneko \etal~\cite{kaneko2018generative} use instead class-level supervision to form the upper level of a hierarchy. The lower-levels are trained with strong inductive bias (e.g. one-hot vectors) and tailored curriculum learning methods. A core difference is that in the related works, they use some form of weak supervision.

The work that is most closely related to our work is that of \cite{lee2020high} which combines VAE and GAN to achieve both high quality of synthesis and unsupervised disentanglement. However, the proposed method requires significant engineering (two-step training process), while it demonstrates only partial disentanglement in high-dimensional distributions. In addition, the latent codes from the latent variable model (VAE) is used as an auxiliary input to the GAN generator. By contrast, in our approach, generation is controlled solely by the learned latent variable model.

\paragraph{Latent variable models}: 
Learning the joint distribution over both observed and latent variables is a crucial topic in machine learning. The normalizing flows~\cite{dinh2014nice, dinh2016density} provide an elegant way to learn the joint distribution; they learn an invertible network. However, their likelihood-based loss results in images that are not competitive to recent generative models. Recently, \cite{khemakhem2019variational} prove that a prerequisite for learning the joint distribution is to have an identifiable model. They make a step further and modify VAE to achieve threefold goals: (i) fit the target distribution, (ii) disentanglement of the variables, (iii) identifiability of the model. They extend VAE, since they argue that it state-of-the-art in (i) and (ii). However, recent works~\cite{nie2020semi} exhibit that GANs outperform VAEs in semi-supervised disentanglement on challenging high-resolution data. This motivates us to use latent variable models in a GAN for realistic generation. To that end, we use moments to fit effective latent variable models for capturing salient features in images.

\paragraph{Self-supervised learning and GAN}: The concept of self-supervised learning has emerged in a GAN training~\cite{chen2018self, huang2020fx, qian2018self}. In \cite{chen2019self} the authors augment the GAN loss with an auxiliary loss to predict rotated versions of the image in the discriminator. In \cite{tran2019self} they propose a self-supervised loss that can stabilize the training, 
while in \cite{arbel2020kale} they involve both the generator and the discriminator for synthesizing samples in the trained model. Self-supervised GANs differ substantially from our goal; they predetermine a set of fixed auxiliary tasks (e.g. rotation), while in our case the goal is to drive the semantic generation through interpretable codes.

 \section{Controlling generation and discovering the latent codes with self-training}
\label{sec:disent_method}
\begin{figure*}[h]
    \centering
    \includegraphics[width=0.85\linewidth]{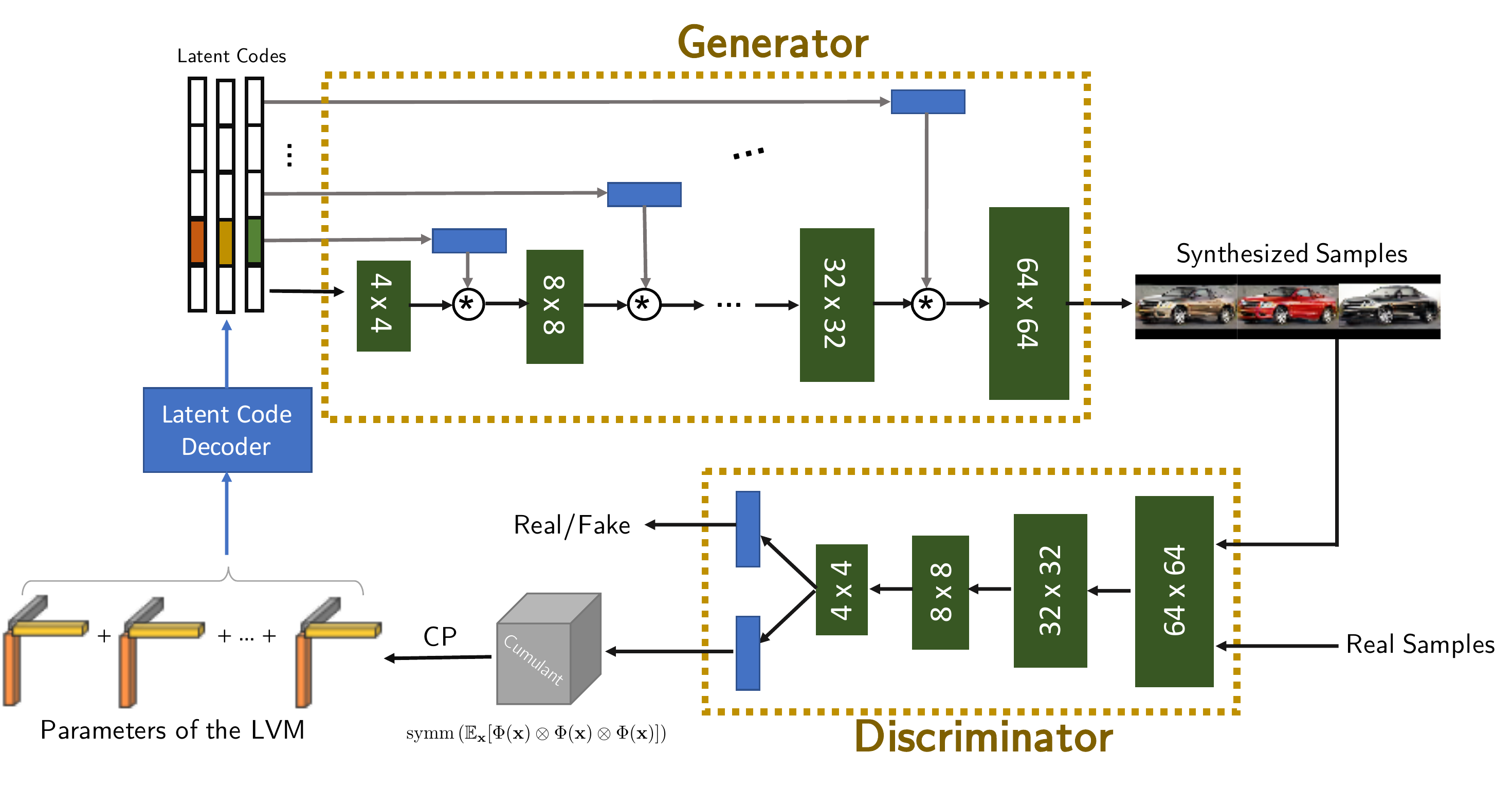}
\caption{Overview of our proposed \modeldisent. The latent variable model (LVM) is learned in the feature space of the discriminator and is used to provide samples in the input of the generator.}
\label{fig:disent_schematic}
\end{figure*}

Our method discovers the disentangled, semantically meaningful latent codes driving the generation in a fully unsupervised manner.
This relies on augmenting a GAN structure towards controllable generation. Specifically, we make the following changes: 
i) instead of sampling from a fixed distribution, 
   the generator takes as input latent codes from a \modeltypedisent, 
ii) a self-training scheme is proposed to discover the latent codes,
iii) a hierarchical structure is used in the generator. 
Each contribution is analyzed below, while in Fig.~\ref{fig:disent_schematic} an abstract schematic of the framework is illustrated. We first review generative adversarial networks, before motivating and describing our proposed method.

\textbf{Notation}: Vectors (scalars) are denoted with boldface (plain) letters, e.g. $\bm{x} (x)$. The outer product is symbolized with $\otimes$.

\subsection{Generative Adversarial Networks}
\label{ssec:disent_method_gan}

\paragraph{Generative Adversarial Networks} A GAN consists of a generator $G$ and a discriminator $D$ engaging in a zero sum game. The goal of the generator is to to model the target distribution $\mathbb{P}_{data}$, while the discriminator aims at discerning the samples synthesized by the generator and the real samples from target (ground-truth) distribution.

The generator samples a latent code from a fixed probability distribution $\mathbb{P}_{\bm{z}}$ (typically Gaussian) and maps it to an image $G(\bm{z})$. 
The discriminator receives both images synthesized by the generator $G(\bm{z})$ 
and samples from the real distribution $\mathbb{P}_{data}$
and tries to distinguish them.
The objective function is: 
\begin{equation}
    \mathcal{L}_{gan} =  \mathbb{E}_{\bm{x}\sim \mathbb{P}_{data}}\Big[\log D(\bm{x})\Big] + \mathbb{E}_{\bm{z}\sim \mathbb{P}_{z}}\Big[\log(1 - D(G(\bm{z}))\Big].
    \label{eq:disent_vanilla_gan_loss}
\end{equation}
This loss is optimized in an alternating manner as $\min_G \max_D \mathcal{L}_{gan}$.  

\paragraph{Style-GAN} proposes hierachical injections in the generator~\cite{karras2018style}.
The injections, i.e. Hadamard products with the inputs, are performed after each layer. In practice, our generator differentiates from Style-GAN injections, as we partition the latent codes (i.e. input to the generator). Specifically, we use $s$ partitions, where $s$ denotes the total number of injections in the generator.
That is, for a latent code $\bm{h} \in \realnum^d$, we use the elements $[h_{1}, \ldots, h_{\frac{d}{s}}]$ in the first injection, the next $\frac{d}{s}$ elements in the second and so on. 
This injection captures higher-order correlations~\cite{chrysos2019polygan}. 
Therefore, by partitioning the latent codes 
we capture the correlations of specific elements in each injection.

\paragraph{Semi-Supervised Style-GAN} augments the training procedure of Style-GAN with two additional losses: a Mixup loss and a consistency loss~\cite{nie2020semi}. 
The consistency loss is expressed as 
$\mathcal{L}_{c} = \left\|\bm{h} - \bm{W}^T v(G(\bm{h}))\right\|_2^2$, where $v$ is the representation from the penultimate layer of the discriminator, $\bm{W}$ a learnable dense layer and $\bm{h}$ is the input to the generator. 
The consistency loss constrains the original and the reconstructed latent codes (i.e., $\bm{h}$) to be close.

The Mixup loss reinforces smoothness in the latent code space. Given a pair of real and fake images $(\bm{I}_r, \bm{I}_f)$ and their corresponding latent codes $(\bm{h}_r, \bm{h}_f)$, we interpolate between the images and the latent codes to obtain $(\bm{I}_s, \bm{h}_s)$. The Mixup loss is then $\mathcal{L}_s = \left\|\bm{h}_s - \bm{W}^T v(\bm{I}_s)\right\|_2^2 $.

In this paper, we build on top of both Style-GAN and its semi-supervised variant and incorporate both losses when training our proposed model.

\subsection{Driving generation with a latent variable model}
In GAN, the generator takes as input a latent code, sampled from a fixed probability distribution.
Typically, a Gaussian or a uniform distribution is selected.
Replacing this fixed distribution with distributions with learnable parameters has been recently explored in the literature, 
both in the context of GAN~\cite{kuznetsov2019prior} 
and other generative models~\cite{tomczak2017vae,bauer2018resampled}.
Simple fixed distributions, e.g. Gaussian, do not encode any semantic meaning such as 3D pose of the object.

Motivated by such works, we replace sampling the latent code from a fixed distribution with latent codes. The latent codes are sampled from a latent variable model.
This underlying model is data-driven and evolves over the training to better reflect the variation of the specific data.
In practice, we use a variant of independent component analysis, which produces latent codes $\bm{h} \sim \mathbb{P}_l$ used by the generator.
The independent component analysis~\cite{comon1994independent} mixes linearly independent signals and then corrupts them with Gaussian noise. However, in this work we normalize the hidden independent signals in the range of $[-1, 1]$. A variant of this model with discrete observations is known as topic model, but we use continuous version and term it as normalized independent component analysis.
If we denote the latent vector  with $\bm{h}$, the Gaussian noise with $\bm{\epsilon}$ and the mixing matrix with $\bm{M}$, the corresponding observation is $\bm{y} = \bm{M}\bm{h} + \bm{\epsilon}$. 
As an ablation, we also explore using a VAE.

\subsection{Learning the latent variable model with self-training}
To obtain the latent codes $\bm{h}$, we use a self-training scheme with a latent variable model (a flexible \modeltypedisent) in the feature space of the discriminator. We learn its parameters through tensor factorization of the higher-order cumulant. We also showcase in ablation that other latent variable models, such as VAEs can also be used as drop-in replacements. 
During inference, the generator samples latent codes $\bm{h} \sim \mathbb{P}_l$ 
from the learnable distribution $\mathbb{P}_l$
and generates samples $G(\bm{h})$. 

\paragraph{Extracting the representation}
We learn the latent variable model (i.e., \modeltypedisent) in the feature space of the discriminator.
Specifically, for an input image $\bm{x}$, the corresponding features $v(\bm{x})$ are extracted, where $v(.)$ corresponds to the penultimate layer of the discriminator. The representation for the latent variable model is computed through a linear layer with learnable vector of parameters $\bm{W}$. 
The final extracted representation is then $\Phi(\bm{x}) = \bm{W}^T v(\bm{x})$.
The mixing matrix is learned through tensor factorization. Specifically, we form the higher-order moments of the features $\Phi(\bm{x})$ (as we elaborate below), and learn their low-rank factorized form.

\paragraph{Learning the parameters of the latent \modeltypedisent{} through tensor factorization}
We assume a \modeltypedisent, which maps from the extracted features to untangled latent codes. 
We propose to learn the parameters of that model by decomposing the higher-order moments. We describe next how in could be derived in the analytical case and how we adapt this in our framework with end-to-end learning.

In the analytical case, we know it is possible to form the moments of the model. From these, the symmetric cumulant tensor (e.g. $\symm\left(\mathbb{E}_{\bm{x}} [\Phi(\bm{x}) \otimes \Phi(\bm{x}) \otimes \Phi(\bm{x})]\right)$ for the third order cumulant) can be obtained by subtracting cross-order terms.
Given a specific latent variable model,
this symmetric tensor -- or cumulant-- can be obtained through a closed-form formula.
Applying low-rank decompositions to it allows the recovery of the parameters of the model~\cite{anandkumar2014tensor}. 

In this paper, all the components are learned end-to-end, including the factors of the decomposition.
Instead of exactly forming the cumulant tensor, following~\citet{arabshahi2016spectral}, we propose to also learn the weights of each of the cross-terms. Specifically, for the second order, this results in $\bm{M_1} + w_1 \bm{M_2}$, where $\bm{M_1} = \mathbb{E}_{\bm{x}} [\Phi(\bm{x}) \otimes \Phi(\bm{x}) ]$ and $\bm{M_2} = \mathbb{E}_{\bm{x}} [\Phi(\bm{x})] \otimes \mathbb{E}_{\bm{x}} [\Phi(\bm{x})]$. Similarly, for the third order term, we form the term $\bm{T_1} + w_2 \bm{T_2} +  w_3 \bm{T_3}+  w_4 \bm{T_4} + w_5 \bm{T_5}$, with:
\begin{align}\nonumber
\centering
   \bm{T_1} &= \mathbb{E}_{\bm{x}} [\Phi(\bm{x}) \otimes \Phi(\bm{x}) \otimes \Phi(\bm{x})], \quad\quad\quad \\ \nonumber
   \bm{T_2} &=  \mathbb{E}_{\bm{x}} [\Phi(\bm{x}) \otimes \Phi(\bm{x})] \otimes \mathbb{E}_{\bm{x}} [\Phi(\bm{x})]\\ \nonumber
   \bm{T_3} &= \mathbb{E}_{\bm{x}} [\Phi(\bm{x}) \otimes \mathbb{E}_{\bm{x}} [\Phi(\bm{x})] \otimes \Phi(\bm{x})] \\ \nonumber
   \bm{T_4} &= \mathbb{E}_{\bm{x}} [\Phi(\bm{x})] \otimes \mathbb{E}_{\bm{x}} [\Phi(\bm{x}) \otimes \Phi(\bm{x})]\\ \nonumber
   \bm{T_5} &= \mathbb{E}_{\bm{x}} [\Phi(\bm{x})] \otimes \mathbb{E}_{\bm{x}} [\Phi(\bm{x})] \otimes \mathbb{E}_{\bm{x}} [\Phi(\bm{x})]
\end{align}

We then factorize the resulting higher-order moments computed from the features $\Phi(x)$ from the discriminator. 
We assume that the cumulants, $\bm{M_1} + w_1 \bm{M_2}$ and $\bm{T}_1 + \sum_{k=1}^5 \bm{T}_k$ admit a rank--$R$ low-rank representation. In other words, we express the (symmetric) cumulants as a weighted sum of $R$ rank--1 tensors.
We learn both the weights of the sum (collected in a vector $\bm{\lambda}$) and the factors $\bm{a}_1, \cdots, \bm{a}_R $ of the decomposition.
Specifically, we minimize the following loss function $\mathcal{L}_{l}$:
\begin{equation}\nonumber
\begin{split}
    \myargmax_{w_k, \bm \lambda, \bm{a}_j} \mathcal{L}_{l} = 
    \left\| 
        \bm{M_1} + w_1 \bm{M_2}
            - \sum_{j=1}^{R} \lambda_j \bm{a}_j \otimes \bm{a}_j
    \right\|_F^2
    + \\
    \left\| 
        \bm{T_1} + \sum_{k=2}^5 w_k \bm{T_k}
            - \sum_{j=1}^{R} \lambda_j \bm{a}_j \otimes \bm{a}_j \otimes \bm{a}_j 
    \right\|_F^2
\end{split}
\end{equation}

The learned parameters $\bm{\lambda}, \bm{a}_j$ form the mixing matrix $\bm{M}$ for the \modeltypedisent. Specifically, the $j^{th}$ row corresponds to the $j^{th}$ factor, i.e. $M_{j, :} = \lambda_j \bm{a}_j$. Then, for a sample $\bm{x} \sim \mathbb{P}_{data}$, we obtain the latent code as $\bm{h} = \bm{M}^{\dagger} (\bm{x} - \epsilon)$ where $\epsilon \sim \mathcal{N}(0, \sigma_{\epsilon}^2\bm{I})$ and $\bm{M}^{\dagger}$ denotes $\bm{M}$'s pseudo-inverse.

To further encourage the disentanglement of the factors, we add an orthogonality regularization term in the loss, i.e. $\mathcal{L}_{l} = \mathcal{L}_{l} + \gamma_o \sum_{j=1}^{R}\left\|\bm{a}_j^T \bm{a}_j - \mathbb{I}\right\|_F^2$ where $\mathbb{I}$ is the identity matrix and $\gamma_o$ is a hyper-parameter.

\textbf{Ablation: Variational Autoencoder} As an ablation, we investigate replacing the higher-order factorization framework with VAEs, i.e., a VAE is learned as an alternative latent variable model. 
Similarly to the \modeltypedisent, we learn the VAE on the feature space of the discriminator. 

Specifically, we maximize the ELBO of the distribution in the feature space of the discriminator: $\mathcal{L}_{l}  =  \mathbb{E}_{q(\bm{h}|\Phi(\bm{x}))}\left[\log p(\Phi(\bm{x}) | \bm{h})\right] - D_{KL}\left( q(\bm{h}|\Phi(\bm{x}))\parallel p(\bm{z}) \right),$ where $\bm{h}$ is the latent code, $q(\bm{h}|\Phi(\bm{x}))$ is the posterior distribution and $D_{KL}$ computes the Kullback-Leibler divergence between two distributions.

\begin{figure*}
    \centering
    \subfloat[\modeldisentmain]{\includegraphics[width=0.49\linewidth]{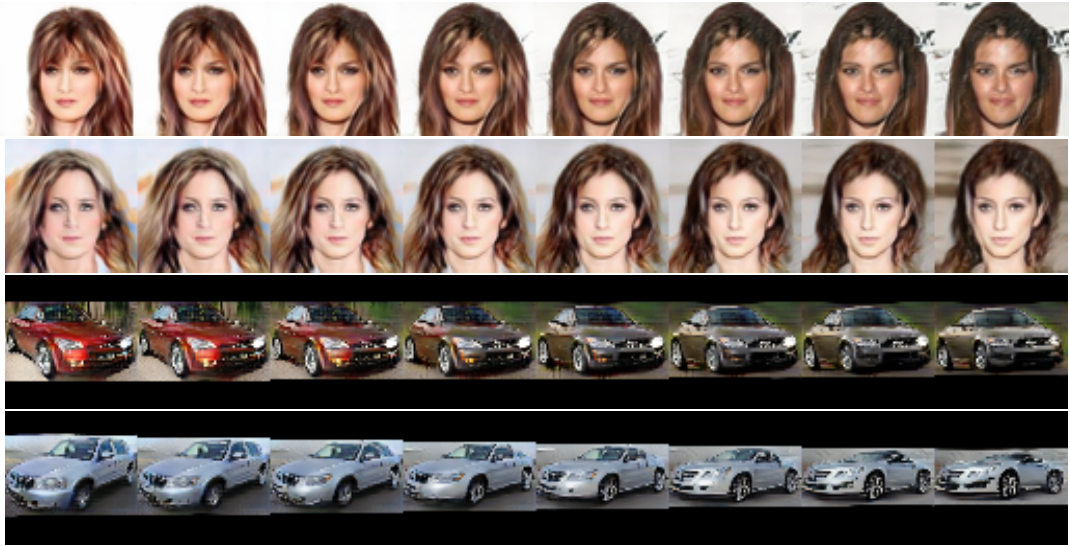}\hspace{1mm}}
    \subfloat[\modeldisent -VAE]{\includegraphics[width=0.49\linewidth]{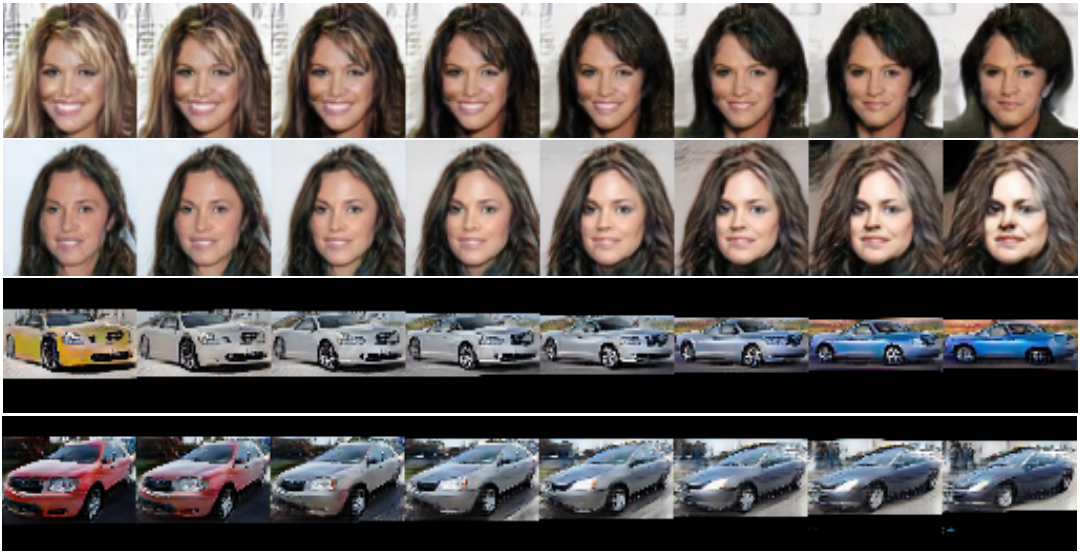}}
    \caption{Linear interpolation from a source to a target latent code. The synthesized images corresponding to each code are visualized. The linear interpolation in the latent space of \modeldisent{} illustrates how the proposed framework can vary all the modes of variation jointly for making nonlinear changes in the image space.}
    \label{fig:disent_linear_interpolation}
\end{figure*}

\begin{table*}[h]
    \caption{Ablation study on the losses.}
    \label{tab:disent_ablation}
    \centering
    \begin{tabular}{@{\extracolsep{\fill}} l r ccc r ccc}
    \toprule
      \multirow{2}{*}[-0.5\dimexpr \aboverulesep + \belowrulesep + \cmidrulewidth]{\textbf{Method}}  & & \multicolumn{3}{c}{\textbf{CelebA}}  & & \multicolumn{3}{c}{\textbf{Cars}} \\
      \cmidrule(l{3pt}r{3pt}){3-6} \cmidrule(l{3pt}r{3pt}){7-9}
      & & \textbf{FID ($\downarrow$)} & \textbf{MAE ($\uparrow$)} & \textbf{LPIPS ($\uparrow$)} & & \textbf{FID ($\downarrow$)} & \textbf{MAE ($\uparrow$)} & \textbf{LPIPS ($\uparrow$)} \\  
      \midrule
            \textbf{\modeldisentmain} & & $\bm{25.21}$ & $\bm{0.265}$ & $\bm{0.040}$       & & $\bm{12.96}$ & ${0.271}$ & ${0.057}$ \\
             \modeldisentmain -$\mathcal{L}_{s}$ & & $32.40$ & $0.249$ & $0.038$  & & $35.23$ & $0.279$ & $0.062$\\
             \modeldisentmain -$\mathcal{L}_c$ & & $33.90$ & $0.247$ & $0.034$    & & $16.66$ & $0.258$ & $0.060$\\
             \modeldisentmain -No ortho & & $30.85$ & $0.250$ & $0.041$           & & $14.81$ & $\bm{0.314}$ & $\bm{0.069}$\\
             \modeldisentmain -$\mathcal{L}_m$ & & $27.47$ & $0.237$ & $0.036$    & & $13.47$ & $0.292$ & $0.060$\\
    \bottomrule
    \end{tabular}
\end{table*}

\paragraph{Training the model}
The self-learning technique of our framework is sensitive to initialization; here, we elaborate on the details of the objective function and the training procedure. For the first $200$ iterations, we use a 'warm-up' of the weights by training a using only the GAN loss. The latent codes in the input of the generator are sampled from a Gaussian distribution. Sequentially, the latent variable model parameters are inserted and the corresponding  $\mathcal{L}_{l}$ loss is added. At $2,000$ iterations, the self-training scheme is added along with the remaining loss terms. 

Our preliminary experiments demonstrated that a soft transition from the prior distribution to the latent distribution is beneficial. To that end, we use a soft transition with an annealing $\kappa$ parameter, i.e. $\bm{h} = \kappa \cdot \bm{h}_l + (1 - \kappa) \cdot \bm{h}_p$ where $\kappa$ starts from $0$ and transitions to $1$ after $8,000$ iterations. The symbol $\bm{h}_l$ denotes a sample from the latent variable model, while $\bm{h}_p$ is a sample from the prior distribution used in the beginning of the training. 

To further induce disentanglement, we add an additional ``masking loss''. The masking loss $\mathcal{L}_m$ encourages each element of the latent code to change one attribute. The masking loss also encourages each change in the image to be `predictable', i.e. different from changes in other attributes. To achieve that, we synthesize a pair of images with a predictable change in their latent codes and try to predict what that change was. That is, given a latent code $\bm{h}$, we duplicate it into $\hat{\bm{h}}$ and sequentially perturb each element of the latent code with uniform noise. The masking loss then tries to predict which element was modified from the $\{\Phi(G(\bm{h})), \Phi(G(\hat{\bm{h}}))\}$ embeddings.

The complete loss function used is 
\begin{align}
    \mathcal{L} = \mathcal{L}_{gan} + \gamma_{l}\mathcal{L}_{l} + \gamma_s\mathcal{L}_{s} + \gamma_c\mathcal{L}_{c}  + \gamma_{m}\mathcal{L}_{m}, 
\end{align}
where $\gamma_{l}, \gamma_s, \gamma_c$ and $\gamma_m$ are regularization hyper-parameters.

 \section{Experiments} 
\label{sec:disent_experiments}

In this section, we describe the experimental setup and the comparisons conducted with the proposed framework. We utilize both the popular CelebA~\citep{liu2015deep} and Cars dataset for our experiments\footnote{CelebA dataset reportedly includes gender and racial biases~\cite{karkkainen2019fairface}, we thus encourage the development of better datasets to address the bias.}. 
CelebA contains $202,000$ images of faces; we use $160,000$ images for training, while the Cars dataset includes $16,000$ images; we use the $12,000$ for training.  
All the images are resized to $64\times64$.

\textbf{Metrics}: The well-established Frechet Inception Distance (FID) metric~\cite{heusel2017gans} is chosen for the generation quality\footnote{The features from the pretrained Inception network of Paszke \etal~\cite{pytorch} are used.}. The metrics proposed for unsupervised disentanglement require an auxiliary encoder to be trained~\cite{higgins2017beta, kim2018disentangling, chen2018isolating}, which is not available in our case. In addition, using direct ground-truth labels is discouraged~\cite{locatello2018challenging}. We opt to report auxiliary semantic metrics that demonstrate the changes affected by each element of the latent code. Specifically, to achieve controllable generation each element of the factor code should modify a factor of variation of the data and be interpretable. To that end, we utilize the following metric: we sample i) an element to perturb and ii) a perturbation in the range $(-1, 1)$. We add the perturbation, generate the two images and then compare them. We repeat this procedure for each element of the latent code for $10$ perturbations. To measure the difference between each pair of images we use both the standard mean absolute error (MAE) and the LPIPS~\cite{zhang2018unreasonable} metric that correlates with the perceptual changes. A higher value in both, means that the single element of the latent code has made a larger (perceptual) change in the image. 

\textbf{Implementation details}: Our implementation is based on the GAN architecture of Miyato \etal~\cite{miyato2018spectral}. That is, both the generator and the discriminator include residual blocks, while the rest hyper-parameters (i.e. optimizer, hinge loss, learning rate) remain unchanged. The injections in the generator follow the implementation of~\cite{chrysos2019polygan}. 
The models are implemented using PyTorch~\cite{pytorch} and TensorLy for all tensor methods~\cite{tensorly}.
We used $\gamma_{l}=1, \gamma_s=0.1, \gamma_c=0.1$. The hyper-parameter $\gamma_m$ is augmented during the training; in iteration $2,000$ it starts as $\gamma_m=1$ and in iteration $10,000$ it takes the value $\gamma_m=100$. The dimensionality of the latent code is $d=50$, i.e., $\bm{h}\in\realnum^{50}$. We apply an element-wise normalization by the max element in the latent codes before feeding them ins the generator. The `masking loss' $\mathcal{L}_m$ is implemented as a cross-entropy loss that predicts which element was modified.

\begin{table*}[h]
\caption{Performance of GAN models in terms of image quality (FID) and controllability (LPIPS, MAE). \modeldisent{} consistently provides better controllability as indicated from the reported metrics of LPIPS and MAE.}
\label{tbl:disent_comparison_main}
    \centering
    \begin{tabular}{l r ccc r ccc}
    \toprule
       \multirow{2}{*}[-0.5\dimexpr \aboverulesep + \belowrulesep + \cmidrulewidth]{\textbf{Method}} & & \multicolumn{3}{c}{\textbf{CelebA}}  & & \multicolumn{3}{c}{\textbf{Cars}} \\
       \cmidrule(l{3pt}r{3pt}){3-6} \cmidrule(l{3pt}r{3pt}){7-9}
       & & \textbf{FID ($\downarrow$)} & \textbf{MAE ($\uparrow$)} & \textbf{LPIPS ($\uparrow$)} & & \textbf{FID ($\downarrow$)} & \textbf{MAE ($\uparrow$)} & \textbf{LPIPS ($\uparrow$)} \\  
       \midrule
             GAN~\cite{miyato2018spectral} & &  $23.21$ & $0.169$ & $0.024$                 & & $23.96$ & $0.243$ & $\bm{0.061}$ \\  
             GAN-Inj & &            $\bm{20.07}$ & $0.179$ & $0.026$                        &  & $15.06$ & $0.240$ & $0.049$ \\  
             GAN-H & &              $21.17$ & $0.178$ & $0.024$                             & & $\bm{12.13}$ & $0.218$ & $0.042$ \\  
             VAE &  &               $91.20$ & $0.003$ & $0.003$                             & & $120.02$ & $0.003$ & $0.003$ \\  
             \textbf{\modeldisentmain} & & $25.21$ & $\bm{0.265}$ & $\bm{0.040}$                   & & $12.96$ & $\bm{0.271}$ & $0.057$ \\ 
             \modeldisent -VAE & &  $23.95$ & $0.166$ & $0.027$                             & & $14.19$ & $0.207$ & $0.041$ \\ 
             \bottomrule 
    \end{tabular}
\end{table*}

\begin{figure*}
    \centering
    \subfloat[\modeldisentmain]{\includegraphics[width=0.49\linewidth]{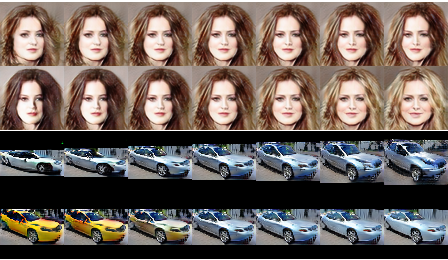}\hspace{1mm}}
    \subfloat[\modeldisent -VAE]{\includegraphics[width=0.49\linewidth]{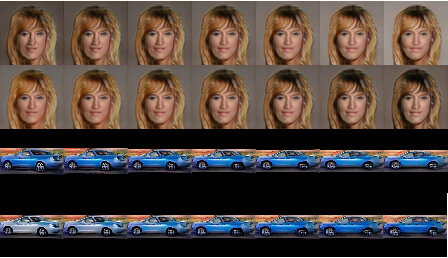}}\\
    \subfloat[SNGAN]{\includegraphics[width=0.49\linewidth]{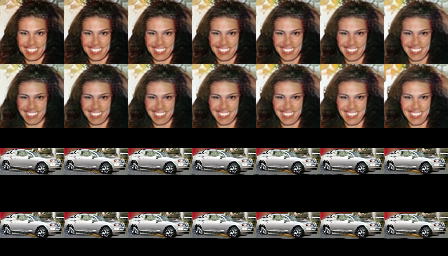}}
    \caption{Each row depicts linear interpolation in a single element of the latent code, while the other elements remain fixed. (a) The first row corresponds to the elements in the first split, while the second row corresponds to an element from the second split. The first row changes the pose and the second the colors (of hair and car respectively). (b) Similar patterns with the VAE model are observed, however the changes are less obvious. (c) One the contrary, in SNGAN, changing a single element does not modify the image.}
    \label{fig:disent_masked_interpolation}
\end{figure*}

\subsection{Ablation studies}
We investigate the different losses as well as the inductive bias of the architecture selected. Specifically, we will denote the model $\textit{\modeldisent-}\mathcal{L}_m$ when it does not include the loss $\mathcal{L}_m$. Similarly for the other losses. The rest of the hyper-parameters are not tuned again, but remain the same. In Table~\ref{tab:disent_ablation}, the quantitative results illustrate that our method performs similarly when removing one loss at a time. However, the final model (with all the losses) outperforms all the variants.

\subsection{Comparison with other models}

Two baseline architectures are considered: i) SNGAN~\cite{miyato2018spectral} (referred to as \emph{GAN} henceforth), ii) VAE with convolutional encoder and decoder. For the SNGAN we consider two additional variants: a) one where we include injections in the generator~\cite{karras2018style} (referred as \emph{GAN-Inj}), b) one with a similar hierarchical generator as in our model (\emph{GAN-H}). The goal is to assess whether the controllable generation is caused by these inductive biases alone. Our model is denoted as `\modeldisentmain', while the ablation with the VAE model in the latent space as `\modeldisent -VAE'. 

The quantitative results in Table~\ref{tbl:disent_comparison_main} exhibit that our method outperforms all the baselines in both the MAE and the LPIPS metric. Intuitively, this means that each element of the factor code makes larger changes to the image than the baselines. To illustrate the differences, we perform two visualizations: a) linear interpolation in the latent space, b) linear interpolation in one element of the latent code at a time. The former visualization in Fig.~\ref{fig:disent_linear_interpolation} simply confirms that changing the latent code linearly results in realistic nonlinear changes in the image. The visualization additionally confirms that \modeldisentmain{} can vary all the modes of variation jointly that result in realistic changes in the image.  
In Fig.~\ref{fig:disent_masked_interpolation}, each row depicts linear interpolation in a single element of the latent code, while the other elements remain fixed. We notice how our model (`\modeldisentmain') changes the pose or the color in any object by using a single element. Both changes bear a semantic interpretation; as we indeed show in the supplementary material consistent changes in the background can be observed. On the contrary, as we show in Fig.~\ref{fig:disent_masked_interpolation} and the supplementary, all the baseline models fail to make such large changes by changing a single element of their latent code.

 \section{Conclusion} 
\label{sec:disent_conclusion}
In this paper, we proposed the first self-trained GAN to enable controllable image generation without any supervision.
Our method discovers disentangled and semantically interpretable factors of variation driving the generation in a fully unsupervised framework. 
The hidden factors are modelled explicitly by a flexible latent variable model, from which the generator samples its inputs. The parameters of that latent variable model are learned through a tensor factorization of higher-order cumulants.
We empirically demonstrate that the codes learned by our model are semantically interpretable and perform favorably when compared with the baseline methods. We expect the discovery of semantically interpretable codes to be useful for multi-modal learning.

\bibliographystyle{unsrt}
\bibliography{egbib}

\end{document}